\documentclass[11pt,a4paper]{article}
\usepackage[hyperref]{acl2021}
\usepackage{times}
\usepackage{latexsym}

\usepackage{graphics}
\usepackage{graphicx}
\usepackage{amsmath}
\usepackage{hyperref}

\usepackage{multirow}
\usepackage{microtype}
\usepackage{colortbl}
\definecolor{Blue2}{RGB}{235, 245, 250}
\usepackage{booktabs}

\usepackage{microtype}

\aclfinalcopy 

\title{ProphetNet-X: Large-Scale Pre-training Models for English, Chinese, Multi-lingual, Dialog, and Code Generation}

\author{Weizhen Qi\textsuperscript{1} \thanks{ \hspace{2mm}Work is done during internship at Microsoft Research Asia.}, Yeyun Gong\textsuperscript{2} \thanks{ \hspace{2mm}Corresponding Author.}  , Yu Yan\textsuperscript{3}, Can Xu\textsuperscript{3}, Bolun Yao\textsuperscript{4}, Bartuer Zhou\textsuperscript{2} \\
\textbf{ Biao Cheng\textsuperscript{2} , Daxin Jiang\textsuperscript{3}, Jiusheng Chen\textsuperscript{3},  Ruofei Zhang\textsuperscript{3}, Houqiang Li\textsuperscript{1}, Nan Duan\textsuperscript{2}}   \\
  \textsuperscript{1}University of Science and Technology of China, \textsuperscript{2}Microsoft Research Asia, \\ \textsuperscript{3}Microsoft, \textsuperscript{4} Nanjing University of Science and Technology \\
  \texttt{\textsuperscript{1}weizhen@mail.ustc.edu.com, lihq@ustc.edu.com}, \\ 
   \texttt{\textsuperscript{2}\{yegong,bazhou,bicheng,nanduan\}@microsoft.com,  }\\  
  \texttt{\textsuperscript{3}\{yyua,caxu,djiang,jiuchen,bzhang\}@microsoft.com \textsuperscript{4}yaobl001@njust.edu.cn} 
  }

\date{}

\begin{document}
\maketitle

\begin{abstract}
Now, the pre-training technique is ubiquitous in natural language processing field. ProphetNet is a pre-training based natural language generation method which shows powerful performance on English text summarization and question generation tasks. In this paper, we extend ProphetNet into other domains and languages, and present the ProphetNet family pre-training models, named ProphetNet-X, where X can be English, Chinese, Multi-lingual, and so on. We pre-train a cross-lingual generation model ProphetNet-Multi, a Chinese generation model ProphetNet-Zh, two open-domain dialog generation models ProphetNet-Dialog-En and ProphetNet-Dialog-Zh. And also, we provide a PLG (Programming Language Generation) model ProphetNet-Code to show the generation performance besides NLG (Natural Language Generation) tasks. In our experiments, ProphetNet-X models achieve new state-of-the-art performance on 10 benchmarks. All the models of ProphetNet-X share the same model structure, which allows users to easily switch between different models. We make the code and models publicly available\footnote{\href{https://github.com/microsoft/ProphetNet}{https://github.com/microsoft/ProphetNet}}, and we will keep updating more pre-training models and finetuning scripts. 
\end{abstract}

\section{Introduction}

\begin{figure*}[ht]
    \centering
    \includegraphics[width = 6.0in]{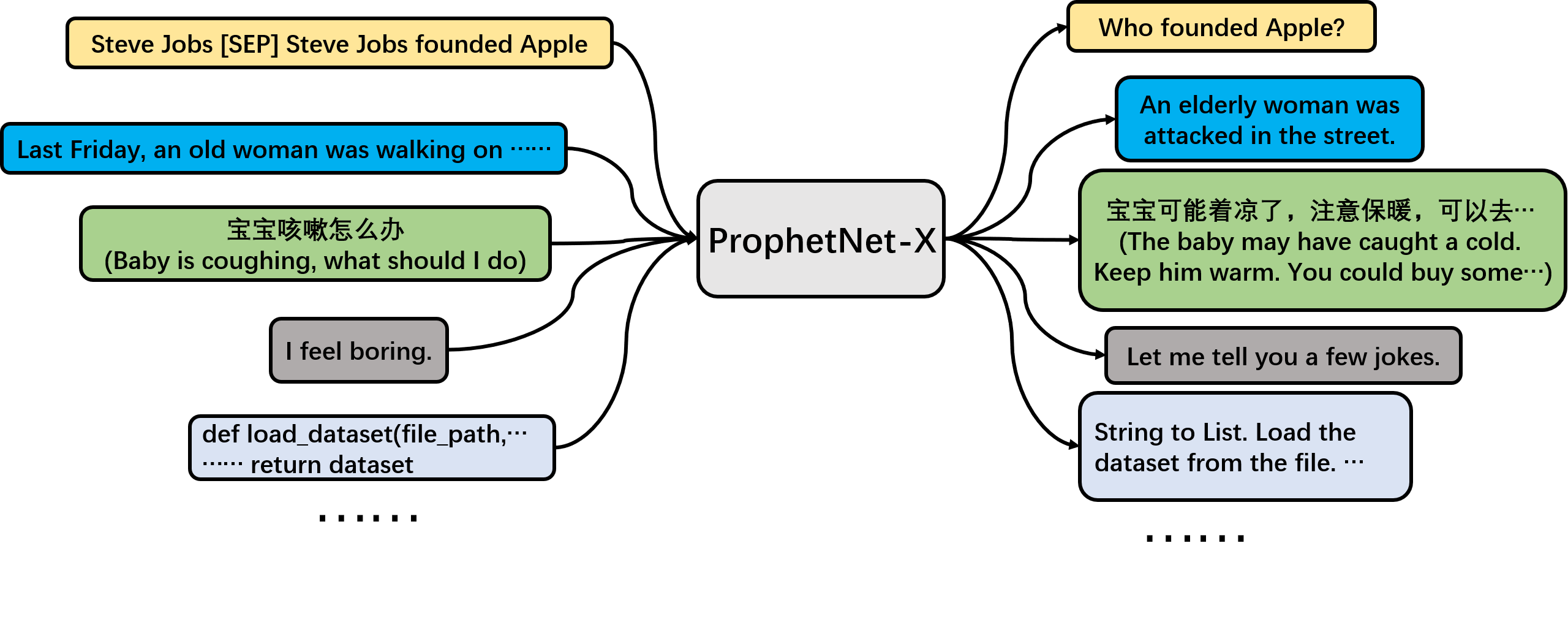}
	\caption{A diagram of ProphetNet-X framework. ProphetNet-X models share the same model structure and cover various languages and domains.} 
	\label{bang.ar}
\end{figure*}

In recent years, quite a few natural language generation pre-training models are proposed~\cite{qi2020prophetnet, lewis2019bart, song2019mass, brown2020language}. Downstream generation tasks benefit from these large scale pre-training models greatly in fluency and accuracy. Researchers also extend these general pre-training works into specific domains such as DialoGPT~\cite{zhang2019dialogpt} is extended from GPT~\cite{brown2020language} for dialog system, mBART~\cite{liu2020multilingual} is extended from BART~\cite{lewis2019bart} for multi-lingual generation, CodeBERT~\cite{feng2020codebert} is extended from BERT~\cite{devlin2018bert} for programming language modeling, etc.

Although there are pre-trained models for some specific domains, it is not convenient for users to find them and set them up. Besides, even some models in the same pre-training family with the same model structure and pre-training tasks, their codes and details vary a lot because of different implementation and backends selection. 

ProphetNet~\cite{qi2020prophetnet} is firstly proposed as an English text pre-training model with future tokens' prediction, and successfully improves the performance on different downstream NLG tasks. 
In this work, we pre-train the ProphetNet on different corpus, respectively. The corpus covers different languages and domains. All the pre-trained models share the same model structure with different vocabularies. We provide six pre-trained models with downstream task finetuning scripts, including ProphetNet-En pre-trained with 160GB English raw text, ProphetNet-Zh pre-trained with 160GB Chinese raw text, ProphetNet-Multi with 101GB Wiki-100 corpus and 1.5TB Common Crawl\footnote{https://commoncrawl.org/} data, ProphetNet-Dialog-En with 60 million sessions Reddit open-domain dialog corpus, ProphetNet-Dialog-Zh with collected Chinese dialog corpus over 30 million sessions, and ProphetNet-Code pre-trained with 10 million codes and documents. ProphetNet-X achieves new state-of-the-art results on 10 benchmarks, including Chinese summarization (MATINF-SUMM~\cite{xu2020matinf} and LCSTS~\cite{hu2015lcsts}), Chinese question answering (MATINF-QA~\cite{xu2020matinf}), cross-lingual generation (XGLUE NTG~\cite{liang2020xglue} and XGLUE QG~\cite{liang2020xglue}), English summarization (MSNews~\cite{liu2020glge}), English dialog generation (DailyDialog~\cite{li2017dailydialog}, PersonaChat~\cite{zhang2018personalizing}, and DSTC7-AVSD~\cite{alamri2019audio}), and code summarization (CodeXGLUE~\cite{lu2021codexglue}). 
Users can simply download the ProphetNet-X repository and find corresponding pre-trained model with downstream task finetuning scripts.

The main contributions of ProphetNet-X can be described as follows:
\begin{itemize}
  \item  We provide a family of pre-trained models named ProphetNet-X, with six models including English and Chinese natural language generation in open-domain and dialog, multi-lingual generation, and code generation.
  \item All the pre-trained ProphetNet-X models share the same model structure. Users only need to simply modify one model file to use it in different language or domain tasks.
  \item We conduct extensive experiments, the results show that ProphetNet-X models achieve new state-of-the-art performance on 10 publicly available benchmarks. 
\end{itemize}

\section{ProphetNet-X}
\subsection{Architecture}
We train different ProphetNet-X models based on ProphetNet. ProphetNet is an encoder-decoder natural language generation model with future n-gram prediction. ProphetNet leverages stacked Transformer encoder layers and stacked multi-stream self-attention Transformer decoder layers. ProphetNet aims to prevent overfitting on strong local correlations such as 2-gram combinations, and deploy future tokens’ prediction to enhance auto-regressive generation ability.

Given the input sequence $x = (x_1, \dots, x_M)$ and output sequence $y = (y_1, \dots, y_T)$, $n$-gram ProphetNet-X replaces the auto-regressive predicting dependency relationship  $p(y_{t}|y_{<t}, x)$ with $p(y_{t:t+n-1}|y_{<t}, x)$. Firstly, ProphetNet-X gets the encoded hidden states with stacked Transformer encoder layers $H_{\rm{enc}} = \textbf{Encoder}(x_1, \dots, x_M)$. Then, decoder with $n$-stream self-attention predicts next $n$ tokens at each time step, as: $p(y_t|y_{<t},x), \dots, p(y_{t+n-1}|y_{<t},x)  = \textbf{Decoder}(y_{<t},H_{\rm{enc}})$. The optimization target of ProphetNet-X can be described as:
\begin{align} 
    \mathcal{L} = &- \sum_{j=0}^{n-1} \alpha_j \cdot \left( \sum_{t=1}^{T-j} \log p_\theta(y_{t+j}|y_{<t},x)\right) \notag\\ 
    = &- \underbrace{\alpha_0 \cdot \left( \sum_{t=1}^T \log p_\theta(y_{t}|y_{<t},x)\right)}_{\text{language modeling loss}} \notag\\  
    &- \underbrace{\sum_{j=1}^{n-1} \alpha_{j} \cdot \left( \sum_{t=1}^{T-j} \log p_\theta(y_{t+j}|y_{<t},x)\right)}_{\text{future n-gram loss}} \nonumber
\end{align}

The details of ProphetNet and multi-stream self-attention can be found in~\citet{qi2020prophetnet}.

\subsection{Pre-training Corpus}

\begin{table*}[h]
\small \centering
\begin{tabular}{c|c|c|c|c|c|c|c|c|c|c|c|c|c} \hline
language & Fr    & It    & Es    & De    & Nl    & Pt    & En    & Sv    & Pl    & Vi    & Ar    & Ru    & Tr    \\ \hline
size(GB) & 77.25 & 74.01 & 72.97 & 71.48 & 71.19 & 71.05 & 68.34 & 67.48 & 67.44 & 67.43 & 65.18 & 64.09 & 62.96 \\ \hline
language &  Ja    & Zh    & Cs    & El    & Ko    & Ro    & Th    & Da    & Bg    & Fi    & Hu    & No    & Hi    \\ \hline
size(GB) & 61.49 & 58.70  & 56.62 & 55.15 & 45.28 & 44.05 & 35.65 & 32.43 & 28.44 & 27.85 & 27.04 & 25.24 & 17.18 \\ \hline
language &  Sk    & Id    & Ca    & Uk    & Lt    & Sr    & Sl    & Hr    & Et    & Lv    & Ka    & Az    & Ur    \\ \hline
size(GB) & 14.78 & 13.68 & 13.08 & 10.80  & 9.20   & 8.59  & 6.86  & 6.51  & 6.47  & 5.48  & 4.16  & 3.38  & 3.13  \\ \hline
language &  Kk    & Ne    & Gl    & My    & Eu    & Gu    & Si    & Ms    & Sq    & Af    & Cy    & Sw    & Bs    \\ \hline
size(GB) & 3.09  & 2.18  & 1.95  & 1.83  & 1.37  & 1.23  & 1.20   & 1.03  & 1.03  & 0.93  & 0.51  & 0.34  & 0.15 \\  \hline 
\end{tabular}
\caption{Statistics of our multi-lingual pre-training corpus. The total pre-training corpus size is 1.54 TB. ISO codes are used to represent each language.} \label{table.multi}
\end{table*}
In this section, we introduce the pre-training corpus for ProphetNet-X.

For ProphetNet-Zh, we collect Chinese Wikipedia, CLUE~\cite{xu2020clue} and Chinese Common Crawl data to reach 160GB. For traditional Chinese data, we firstly use OpenCC~\footnote{https://github.com/BYVoid/OpenCC} to convert them to simplified Chinese. The pre-training corpus includes common webs, online forums, comments websites, Q\&A websites, Chinese Wikipedia, and other encyclopedia websites. We build a simplified Chinese char vocabulary. The char vocabulary size is 9,360.

For ProphetNet-Multi, besides Wiki-100 corpus, we select 52 common languages to collect and clean multi-lingual data from Common Crawl. After cleaning and tokenizing, the Common Crawl corpus size we use is described in Table~\ref{table.multi}. The ProphetNet-Multi vocabulary is same as XLM-R~\cite{conneau2019unsupervised} 250k sentencepiece\footnote{https://github.com/google/sentencepiece} model.

For ProphetNet-Dialog-En, we utilize Reddit comments dataset \cite{zhou2018commonsense, galley2019grounded}. We firstly load the weights of ProphetNet-En then clean 60 million sessions for pre-training.

For ProphetNet-Dialog-Zh, we use the pre-training corpus from ~\citet{wang2020large} and we crawled 18.2 million dyadic dialogues (conversation between two persons) longer than or equal to 2 turns (one turn denotes one utterance from one person) from the Douban group\footnote{https://www.douban.com/group} which is a popular social networking service in China. The pre-training corpus size comparison between ~\citet{wang2020large} and ProphetNet-Dialog-Zh is shown in Table~\ref{table.dialog.zh}. We also load the pre-trained model from ProphetNet-Zh before pre-training, which already contains external knowledge from open-domain Chinese corpus. 

\begin{table}[h]
\small \centering
\begin{tabular}{c|c|c} \hline
Corpus Size          & Single-turn & Multi-turn \\ \hline
LCCC-base            & 3,354,382   & 3,466,607  \\
LCCC-large           & 7,273,804   & 4,733,955  \\ \hline
\small{ProphetNet-Dialog-Zh} &  23,309,502     &  6,985,425  \\ \hline
\end{tabular}
\caption{Statistics of Chinese Dialog pre-training corpus} \label{table.dialog.zh}
\end{table}

For ProphetNet-Code, we conduct pre-training on both PLs (Programming Languages) and their describing NL (Natural Language). We use the pre-training corpus provided by CodeSearchNet~\cite{husain2019codesearchnet}. It covers 6 programming languages, including Go, Java, Javascript, PHP, Python, and Ruby. We employ the same sentencepiece tokenizer as CodeBERT~\cite{feng2020codebert}. The tokenizer is used for both PL and NL, with a vocabulary size 50,365.

For ProphetNet-En, we directly take the model pre-trained in ProphetNet~\cite{qi2020prophetnet}. It is pre-trained with 160GB English raw texts, including Wikipedia, books, stories, news, and web texts. The vocabulary of ProphetNet-En is same as BERT sub-words vocabulary. The vocabulary is based on bpe subwords with a max length matching algorithm. Its vocabulary size is 30,522.

\section{Experiments}

\begin{table*}[h]
\small \centering
\begin{tabular}{l|ccc|ccc|ccc}  \hline
\multirow{2}{*}{Method} & \multicolumn{3}{c|}{MATINF-QA} & \multicolumn{3}{c|}{MATINF-SUMM} & \multicolumn{3}{c}{LCSTS} \\
                        & R-1      & R-2     & R-L      & R-1       & R-2      & R-L      & R-1     & R-2    & R-L    \\ \hline
TextRank \cite{mihalcea2004textrank} & -        & -       & -        & 35.53    & 25.78    & 36.84    & 24.38   & 11.97  & 16.76  \\
LexRank \cite{erkan2004lexrank}  & -        & -       & -        & 33.08     & 23.31    & 34.96    & 22.15   & 10.14  & 14.65  \\
Seq2Seq \cite{sutskever2014sequence}                 & 16.62    & 4.53    & 10.37    & 23.05     & 11.44    & 19.55    & -       & -      & -      \\
Seq2Seq+Att \cite{luong2015effective}             & 19.62    & 5.87    & 13.34    & 43.05     & 28.03    & 38.58    & 33.80   & 23.10  & 32.50  \\
WEAN  \cite{ma2018query}                  & -        & -       & -        & 34.63     & 22.56    & 28.92    & 37.80   & 25.60  & 35.20  \\
Global Encoding \cite{lin2018global}        & -        & -       & -        & 49.28     & 34.14    & 47.64    & 39.40   & 26.90  & 36.50  \\
BertAbs  \cite{liu2019text}               & -        & -       & -        & 57.31     & 44.05    & \textbf{55.93}    & -       & -      & -      \\
MTF-S2S$_{single}$ \cite{xu2020matinf}   & 20.28    & 5.94    & 13.52    & 43.02     & 28.05    & 38.55    & 33.75   & 23.20  & 32.51  \\
MTF-S2S$_{multi}$    \cite{xu2020matinf}             & 21.66    & 6.58    & 14.26    & 48.59     & 35.69    & 43.28    & -       & -      & -      \\ \hline
ProphetNet-Zh           &  \textbf{24.18}   & \textbf{6.38}   & \textbf{15.47} &    \textbf{58.82}    &    \textbf{44.96}      &   54.26       &   \textbf{42.32}   & \textbf{27.33}   &  \textbf{37.08} \\  
\hline
\end{tabular}
\caption{Results of ProphetNet-Zh on MATINF-QA, MATINF-SUMM, and LCSTS. ``R-1'', ``R-2'', and ``R-L'' represent ``ROUGE-1'', ``ROUGE-2'', and ``ROUGE-L'', respectively.}\label{tab.result.zh}
\end{table*}

\begin{table*}[!h]
\small
\centering
\begin{tabular}{c|l|ccccccc|c}  \hline
Task                                    & Model            & De  & En   & Es   & Fr  & It   & Pt  & Ru  & AVG  \\  \hline
\multirow{5}{*}{QG}    & M-BERT \cite{devlin2018bert}          & 0.1 & 7.8  & 0.1  & 0.1 & 0.2  & 0.1 & -   & 1.4  \\
                 & XLM-R$_{base}$ \cite{conneau2019unsupervised}      & 0.1 & 6.0  & 0.0  & 0.0 & 0.1  & 0.0 & -   & 1.0  \\
                 & Unicoder$_{DAE}$ \cite{liang2020xglue}     & 3.0 & 14.0 & 12.4 & 4.2 & 15.8 & 8.3 & -   & 9.6  \\
                 & Unicoder$_{FNP}$  \cite{liang2020xglue}   & 3.7 & 13.9 & 14.8 & 4.9 & 17.0 & 9.5 & -   & 10.6 \\ 
                 & ProphetNet-Multi &  \textbf{4.9}  &  \textbf{14.9}  & \textbf{17.0}  & \textbf{6.0} & \textbf{19.2}    & \textbf{11.3}  & -   &  \textbf{12.2}  \\  \hline \hline
\multirow{5}{*}{NTG}                    & M-BERT    \cite{devlin2018bert}         & 0.7 & 9.0  & 0.4  & 0.4 & -    & -   & 0.0 & 2.1  \\
                                        & XLM-R$_{base}$   \cite{conneau2019unsupervised}      & 0.6 & 8.1  & 0.4  & 0.3 & -    & -   & 0.0 & 1.9  \\
                                        & Unicoder$_{DAE}$    \cite{liang2020xglue}   & 6.8 & 15.6 & 9.0  & 8.7 & -    & -   & 7.7 & 9.6  \\
                                        & Unicoder$_{FNP}$  \cite{liang2020xglue}    & 7.5 & 15.8 & 11.9 & 9.9 & -    & -   & 8.4 & 10.7 \\ 
                                        & ProphetNet-Multi & \textbf{8.7}  &  \textbf{16.7}  & \textbf{12.7} & \textbf{11.4}   & -  & -   & \textbf{8.5} &  \textbf{11.6}  \\  \hline
\end{tabular}
\caption{Results of ProphetNet-Multi on XGLUE zero-shot cross-lingual generation task. Task QG and NTG represent Question Generation and News Title Generation. Numbers in this table are BLEU-4 scores.}
\label{tab:prophetnet-multi}
\end{table*}

\begin{table*}[h]
\small
\begin{tabular}{l|ccccccc}
\hline
Model      & BLEU-1 & BLEU-2 & BLEU-3 & BLEU-4 & METEOR & ROUGE-L & CIDEr \\ \hline
AVSD Baseline \cite{alamri2019audio}  & 0.629  & 0485   & 0.383  & 0.309  & 0.215  & 0.487  & 0.746 \\
CMU Sinbad’s  \cite{sanabria2019cmu}      & 0.718  & 0.584  & 0.478  & 0.394  & 0.267  & 0.563  & 1.094 \\
PLATO  \cite{bao-etal-2020-plato}    & 0.784  & 0.637  & 0.525  & 0.435  & 0.286  & 0.596  & 1.209 \\  \hline
ProphetNet-Dialog-En & \textbf{0.823}  & \textbf{0.688}  & \textbf{0.578} & \textbf{0.482}  & \textbf{0.309} & \textbf{0.631} & \textbf{1.354}  \\
 \hline
\end{tabular}
\caption{Results of ProphetNet-Dialog-En on DSTC7-AVSD. }
\label{tab:dstc7-avsd}
\end{table*}

\begin{table*}[h]
\centering
\label{tab:dailidialog}
\small
\begin{tabular}{l|ccccc|ccccc}
\hline
\multicolumn{1}{c|}{\multirow{2}{*}{Model}} & \multicolumn{5}{c|}{DailyDialog}           & \multicolumn{5}{c}{PersonaChat}          \\
\multicolumn{1}{c|}{}                       & B-1 & B-2 & D-1 & D-2 &AVG& B-1 & B-2 & D-1 & D-2&AVG \\ \hline
Seq2Seq~\cite{vinyals2015neural}            & 0.336  & 0.238  & 0.03       & 0.128 & 0.183 & 0.448  & 0.353  & 0.004      & 0.016 &  0.205  \\
iVAE\_MI~\cite{fang2019implicit}            & 0.309  & 0.249  & 0.029      & 0.25  & 0.209     & -      & -      & -          & -      & -    \\
LIC~\cite{golovanov2019large}                                         & -      & -      & -          & -    &-      & 0.405  & 0.320   & 0.019      & 0.113 & 0.214     \\
PLATO w/o latent  \cite{bao-etal-2020-plato}& 0.405  & 0.322  & 0.046      & 0.246   &0.255   & 0.458  & 0.357  & 0.012      & 0.064   &0.223   \\
PLATO \cite{bao-etal-2020-plato}            & 0.397  & 0.311  & \textbf{0.053}      & \textbf{0.291}   &0.263   & 0.406  & 0.315  & \textbf{0.021}      & \textbf{0.121}   &0.216   \\  \hline
ProphetNet-Dialog-En       &  \textbf{0.461} & \textbf{0.402}   &  0.038    &   0.208 & \textbf{0.277}   & \textbf{0.459}  & \textbf{0.382} &  0.010 & 0.060  &\textbf{0.228}  \\ \hline
\end{tabular}
\caption{Results of ProphetNet-Dialog-En on DailyDialog and PersonaChat. ``B-1'', ``B-2'', ``D-1'' and ``D-2'' represent ``BLEU-1'', ``BLEU-2'', ``Distinct-1'' and `` Distinct-2'', respectively.}  \label{tab.daily.personachat}
\end{table*}

\subsection{Pre-training Settings}
We carry out pre-training with 12-layer encoder, 12-layer decoder ProphetNet models. The hidden size is 1,024, feed forward size is 4,096, future tokens' prediction length is 2. Both the max sequence lengths of the input and output are set to 512.

For ProphetNet-En, ProphetNet-Zh, ProphetNet-Multi, ProphetNet-Dialog-En, and ProphetNet-Code, we carry out un-supervised pre-training with masked span prediction task. Spans of continuous tokens are masked out from the encoder input sentences and predicted from the decoder side. We masked continuous 9 tokens in every 64 tokens from the encoder side, and predict the 9 tokens on the decoder side. In other words, for maximum 512 encoder sequence length, totally $8(spans) \times 9(tokens\ per\ span) = 72$ tokens are masked and predicted. If the last part does not reach a maximum length of 64, 15\% continuous tokens are masked. ProphetNet-Dialog-En has special tokens [X\_SEP] to separate turns in a session and [SEP] to separate different sessions. For ProphetNet-Dialog-Zh, we conduct supervised pre-training. Previous turns of dialogs are fed into the encoder, and the response is predicted from the decoder. It means that for a multi-turn session with $n$ sentences, $n-1$ samples are created for pre-training. The pre-trained ProphetNet-Dialog-Zh can be used to directly generate dialogs without finetuning. 

We carry out pre-training on NVIDIA Tesla V100 GPUs, and the total cost exceeds 30,000 GPU hours.

\subsection{Finetuning Benchmarks}

For different ProphetNet-X models, we select different benchmarks to evaluate them, respectively. 

For ProphetNet-Zh, we evaluate our pre-trained model with MATINF-QA~\cite{xu2020matinf} for generative question answering task,  MATINF-SUMM~\cite{xu2020matinf} and LCSTS~\cite{hu2015lcsts} for summarization task.

For ProphetNet-Multi, we follow Unicoder$_{FNP}$ to evaluate on XGLUE~\cite{liang2020xglue} for cross-lingual zero-shot generation tasks. The pre-trained multi-lingual model is finetuned with English supervised data and inference with English and other un-seen languages data. There are NTG (News Title Generation) and QG (Question Generation) tasks. 

For ProphetNet-Dialog-En, we carry out finetuning on DailyDialog \cite{li2017dailydialog} for chit-chat generation, Persona-Chat \cite{zhang2018personalizing} for knowledge grounded conversation generation and DSTC7-AVSD \cite{alamri2019audio} for conversational question answering.

For ProphetNet-Dialog-Zh, we use the STC~\cite{shang2015neural} single-turn open-domain dialog dataset cleaned by~\citet{wang2020large}, and real-world Xiaoice Chinese dialog dataset for evaluation.

For ProphetNet-Code, we evaluate the performance on code summarization task from CodeXGLUE~\cite{lu2021codexglue}.

For ProphetNet-En, we reports the results on summarization tasks CNN/DM~\citep{hermann2015cnndm}, Gigaword~\citep{rush2015neural}, and MSNews~\cite{liu2020glge}; question generation tasks SQuAD 1.1 \citep{rajpurkar2016squad} and MSQG~\cite{liu2020glge}.

\subsection{Results}

For ProphetNet-Zh, we see significant improvements in Table ~\ref{tab.result.zh}. TextRank \cite{mihalcea2004textrank} and LexRank \cite{erkan2004lexrank}  are extractive baselines and others are abstractive baselines. MTF-S2S$_{single}$ \cite{xu2020matinf}  and MTF-S2S$_{multi}$ denote single task finetuning and multi-task finetuning on MATINF dataset. We see consistent gains on both Chinese question answering task and summarization tasks.

For ProphetNet-Multi, we show the results in Table~\ref{tab:prophetnet-multi}, Unicoder$_{DAE}$ and Unicoder$_{FNP}$ are pre-trained on Wiki-100 with denoising auto encoder task and ProphetNet, respectively. Comparing the results between the Unicoder$_{FNP}$ and ProphetNet-Multi, we see that more pre-training corpus improves supervised English inference results and other zero-shot languages inference performance. And compared with other baseline methods, ProphetNet-Multi achieves new state-of-the-art results on both NTG and QG tasks.  

For English open-domain dialog generation, we show the results in Table~\ref{tab:dstc7-avsd} and Table~\ref{tab.daily.personachat}, compared with strong new proposed PLATO~\cite{bao-etal-2020-plato}, we see that ProphetNet-Dialog achieves performance improvements. 
\begin{table}[h]
\small
\begin{tabular}{l|cc}  \hline
Models                & B-2 & B-4   \\  \hline
Seq2Seq-Attn  \cite{luong2015effective}            & 3.93   & 0.9                         \\
Transformer \cite{vaswani2017attention}          & 6.72   & 3.14                         \\
GPT$_{Novel}$ \cite{wang2020large}          & 5.96   & 2.71                     \\
CDialGPT$_{LCCC-base}$ \cite{wang2020large}  & 6.48   & 3.08                    \\
CDialGPT2$_{LCCC-base}$ \cite{wang2020large} & 5.69   & 2.50                       \\
CDialGPT$_{LCCC-large}$ \cite{wang2020large} & 6.63   & \textbf{3.20}                     \\ \hline
ProphetNet-Dialog-Zh w/o finetuning        &  2.54   & 0.75     \\ 
ProphetNet-Dialog-Zh  w/ finetuning     &  \textbf{6.78}    & 3.05     \\ \hline                     
\end{tabular}
\caption{Results of ProphetNet-Dialog-Zh on STC dataset. ``B-2'', and ``B-4'' represent ``BLEU-2'' and ``BLEU-4'', respectively.}\label{tab.dialog.zh.stc}
\end{table}

\begin{table}[h]
\small \centering
\begin{tabular}{l|ccc|c}  \hline
Setting                & Win & Lose & Tie & Kappa   \\  \hline
Ours-C vs Xiaoice-C          & 68\%   & 26\% & 6\% &  0.73 \\
Ours-C vs Xiaoice-S         & 76\%   & 24\% & 0\%   &  0.65    \\
Ours-S vs Xiaoice-S         & 81\%   & 19\% & 0\%    &  0.67 \\ \hline                     
\end{tabular}
\caption{Human evaluated results for ProphetNet-Dialog-Zh on real-world Xiaoice dataset. Here, Ours means ProphetNet-Dialog-Zh, Xiaoice means old Xiaoice retrieval based dialog system. -S(single-turn) denotes only the last turn is fed to our model or Xiaoice traditional single-turn retrieval model. -C(context) denotes feeding dialog history into our model or Xiaoice traditional multi-turn retrieval model.  }\label{tab.dialog.zh.xiaoice}
\end{table}

\begin{table*}[h]
\small \centering
\begin{tabular}{l|ccccccc}  \hline
Models      & Ruby  & Javascript & Go    & Python & Java  & PHP   & overall \\  \hline
Seq2Seq~\cite{vinyals2015neural}            & 9.64  & 10.21      & 13.98 & 15.93  & 15.09 & 21.08 & 14.32   \\
Transformer   \cite{vaswani2017attention}      & 11.18 & 11.59      & 16.38 & 15.81  & 16.26 & 22.12 & 15.56   \\
RoBERTa \cite{liu2019roberta}    & 11.17 & 11.90       & 17.72 & 18.14  & 16.47 & 24.02 & 16.57   \\
CodeBERT \cite{feng2020codebert} & 12.16 & 14.90       & 18.07 & 19.06  & 17.65 & 25.16 & 17.83   \\
PLBART \cite{ahmad2021unified} & 14.11 & 15.56   & \textbf{18.91} & \textbf{19.30}  & 18.45 & 23.58 & 18.32  \\
\hline
Prophetnet-Code & \textbf{14.37} & \textbf{16.60}       & 18.43 & 17.87  & \textbf{19.39} & \textbf{24.57}  & \textbf{18.54}   \\ \hline
\end{tabular}
\caption{Results of ProphetNet-Code on CodeXGLUE for code-to-text summarization task. Numbers in this table are smoothed BLEU-4 scores.}\label{tab.code.summ}
\end{table*}

\begin{table*}[h]
\small \centering
\begin{tabular}{l|ccc|ccc|ccc}  \hline
\multirow{2}{*}{Method} & \multicolumn{3}{c|}{CNN/DM} & \multicolumn{3}{c|}{Gigaword} & \multicolumn{3}{c}{MSNews} \\
                        & R-1      & R-2     & R-L      & R-1       & R-2      & R-L      & R-1     & R-2    & R-L    \\ \hline
LSTM \cite{bahdanau2014neural} & 37.3&15.7&34.4 & 33.6&15.4&31.2  & 30.0&14.6&27.7   \\
Transformer \cite{vaswani2017attention} & 39.5&16.7&36.7 & 36.4&17.7&33.8 & 33.0&15.4&30.0   \\
MASS \cite{song2019mass} & 42.9&19.8&39.8 & 38.9&20.2&36.2 & 40.4&21.5&36.8     \\ 
BART \cite{lewis2019bart} & 44.1&\textbf{21.2}&40.9 & 37.5&17.6&34.3  & 43.8&24.0&39.2    \\ \hline
ProphetNet-En  & \textbf{44.2}&21.1&\textbf{41.3} & \textbf{39.5}&\textbf{20.4}&\textbf{36.6} &  \textbf{44.1}&\textbf{24.4}&\textbf{40.2}    \\   \hline
\end{tabular}%
\caption{Results of ProphetNet-En for text summarization. ``R-1'', ``R-2'', and ``R-L'' represent ``ROUGE-1'', ``ROUGE-2'', and ``ROUGE-L'', respectively.}\label{tab.result.summ}
\end{table*}

Results for ProphetNet-Dialog-Zh on STC can be seen in Table \ref{tab.dialog.zh.stc}. In addition, Table~\ref{tab.dialog.zh.xiaoice} shows the results on real-world Xiaoice dialog dataset with human evaluation. Results in Table~\ref{tab.dialog.zh.stc} hint that for dialog generation, the auto-evaluation metrics (BLEU-2 and BLEU-4) may fail because open-domain dialog outputs could be very different from the given golden targets but still good responses. We observe that ProphetNet-Dialog-Zh without finetuning can generate fluent and meaningful responses but have lower BLEU scores because of the writing style difference. Thus, we conduct a human evaluation as in ~\cite{zhao2020learning}. We randomly collect 500 single-turn and 500 multi-turn context-response pairs from the online logs of the real-word dialog system Xiaoice. Then, we recruit $3$ native speakers as human annotators. The annotators have to judge which response is better, based on informativeness, consistency, and fluency of the responses. If an annotator cannot tell which response is better, he/she is required to label a ``Tie''. With the experts' annotation, we see that ProphetNet-Dialog-Zh obviously outperforms Xiaoice retrieval based old system. Kappa ~\cite{fleiss1973equivalence} values of all models exceed 0.6, indicating substantial agreement overall annotators.

For ProphetNet-Code, the code summarization results are shown in Table \ref{tab.code.summ}. We can see new state-of-the-art results are obtained with ProphetNet-Code. It shows that ProphetNet-X models not only benefit from pre-training on natural language generation tasks but also perform well in programming language tasks. 

\begin{table}[h]
\small
\setlength{\tabcolsep}{1.5mm}{
\begin{tabular}{l|ccc|ccc}
\hline
\multicolumn{1}{c|}{\multirow{2}{*}{Model}} & \multicolumn{3}{c|}{SQuAD 1.1}    & \multicolumn{3}{c}{MSQG}          \\
\multicolumn{1}{c|}{}                       & R-L & B-4 & MTR  & R-L & B-4 & MTR  \\ \hline
LSTM   & 27.2 & 3.8 & 8.9 & 25.3 & 3.5 & 14.1  \\
Transformer   &   30.7 & 4.8 & 10.9 & 29.3 & 5.1 & 16.6   \\
MASS & 49.9 & 21.3 & 25.2 & \textbf{38.9} & 9.5 & 23.5   \\ 
BART  &   50.3 & 22.0 & \textbf{26.4} & 38.8 & 9.2 & \textbf{24.3}   \\ \hline
ProphetNet-En  &  \textbf{51.5} & \textbf{22.5} & 26.0 & 38.3 & \textbf{9.6} & 23.3   \\   \hline
\end{tabular}%
}
\caption{Results of ProphetNet-En for question generation on SQuAD1.1 and MSQG. ``R-L'', ``B-4'', and ``MTR'' represent ``ROUGE-L'', ``BLEU-4'', and ``METEOR'', respectively.}\label{tab.result.qg}. 
\end{table}

For ProphetNet-En, we report the results for ProphetNet in Table~\ref{tab.result.summ} and Table~\ref{tab.result.qg}. We also report the results for two new tasks MSNTG and MSQG introduced from GLGE~\cite{liu2020glge}.

\section{Related Work}
ProphetNet~\cite{qi2020prophetnet} is the most related to our work since we carry out pre-training based on it. 
Other related works involve pre-training works in different domains. For English generation pre-training, MASS~\cite{song2019mass} proposes an unsupervised pre-training task with span masked and recover. 
BART~\cite{lewis2019bart} feeds corrupted sentences into the encoder and reconstructs the original sentences. GPT~\cite{radford2019language} models perform language modeling pre-training with Transformer decoder. 
For multi-lingual pre-training, mBART~\cite{liu2020multilingual} introduces language labels to adopt BART denoising pre-training. Based on GPT~\cite{radford2019language}, DialoGPT~\cite{zhang2019dialogpt} and CDialGPT \cite{wang2020large} adopts language model pre-training with English and Chinese dialog corpus respectively. CodeBERT~\cite{feng2020codebert} and GraphCodeBERT~\cite{guo2020graphcodebert} are two pre-training models for programming languages. PLBART  \cite{ahmad2021unified} is similar to multi-lingual BART with language tags to perform denoising pre-training on programming languages.

\section{Conclusion}
In this paper, we pre-train ProphetNet-X on various languages and domains, including open-domain (for English, Chinese, and Multi-lingual), dialog (for English and Chinese), and programming (for Ruby, Javascript, Go, Python, Java, and PHP). All the models share the same model structure and are easy to use. Extensive experiments show that ProphetNet-X achieves new state-of-the-art performance on 10 benchmarks. In the future, we will extend ProphetNet-X to support more domains such as biomedical text and protein pre-training.    

\bibliographystyle{acl_natbib}
\bibliography{anthology,acl2021}


\end{document}